\pdfoutput=1

\documentclass[hidelinks,12pt]{article}

\usepackage{hyperref, url} 
\usepackage{graphicx,amsfonts,psfrag,layout,subcaption,array,longtable,lscape,booktabs,dcolumn,amsmath,amssymb,amssymb,amsthm,setspace,epigraph,chronology,color,colortbl,wasysym,diagbox,natbib,colortbl,authblk,commath,upgreek,threeparttable,adjustbox,textcomp}
\usepackage[]{graphicx}\usepackage[]{color}
\usepackage[page]{appendix}
\usepackage[section]{placeins}
\usepackage[linewidth=1pt]{mdframed}
\usepackage[margin={1in}]{geometry} 

\title{RNN-based counterfactual prediction, with an application to homestead policy and public schooling} 
\author[1]{Jason Poulos\thanks{\emph{Address for correspondence:} 180 Longwood Avenue, Boston, MA 02115. \emph{E-mail:} poulos@hcp.med.harvard.edu. Poulos acknowledges the support of the National Science Foundation Graduate Research Fellowship (DGE-1106400) and the National Science Foundation under Grant DMS-1638521 to the Statistical and Applied Mathematical Sciences Institute. This work used the Extreme Science and Engineering Discovery Environment (XSEDE) Comet GPU at the San Diego Supercomputer Center through allocation SES180010. Code to reproduce the results of the paper is available at \url{https://github.com/jvpoulos/rnns-causal}.}}
\author[2]{Shuxi Zeng}
\affil[1]{Department of Health Care Policy, Harvard Medical School}
\affil[2]{Department of Statistical Science, Duke University}
\date{}
\usepackage[autosize]{dot2texi}
\usepackage{tikz}
\usetikzlibrary{shapes,arrows}

\usepackage{xr}
\usepackage[bottom,multiple]{footmisc}

\usepackage{mathtools,tikz,caption}
\DeclareRobustCommand\sampleline[1]{%
	\tikz\draw[#1] (0,0) (0,\the\dimexpr\fontdimen22\textfont2\relax)
	-- (2em,\the\dimexpr\fontdimen22\textfont2\relax);%
}

\usetikzlibrary{plotmarks}

\usepackage{multicol}

\usepackage{xcolor}

\definecolor{Darjeeling11}{HTML}{FF0000}
\definecolor{Darjeeling15}{HTML}{5BBCD6}

\DefineFNsymbols{mySymbols}{{\ensuremath\dagger}{\ensuremath\ddagger}\S\P
	*{**}{\ensuremath{\dagger\dagger}}{\ensuremath{\ddagger\ddagger}}}
\setfnsymbol{mySymbols}

\renewcommand\footnotemark{}

\usepackage{tabularx}
\newcolumntype{Y}{>{\raggedleft\arraybackslash}X}

\definecolor{Gray}{gray}{0.9}

\newcommand{\captionfonts}{\normalsize}

\makeatletter  
\long\def\@makecaption#1#2{%
	\vskip\abovecaptionskip
	\sbox\@tempboxa{{\captionfonts #1: #2}}%
	\ifdim \wd\@tempboxa >\hsize
	{\captionfonts #1: #2\par}
	\else
	\hbox to\hsize{\hfil\box\@tempboxa\hfil}%
	\fi
	\vskip\belowcaptionskip}

\newtheorem*{assumption*}{\assumptionnumber}
\providecommand{\assumptionnumber}{}
\makeatletter
\newenvironment{assumption}[2]
{%
	\renewcommand{\assumptionnumber}{Assumption #1}%
	\begin{assumption*}%
		\protected@edef\@currentlabel{#1}%
	}
	{%
	\end{assumption*}
}
\makeatother

\newcommand{\E}{\mathrm{E}}

\newcommand\independent{\protect\mathpalette{\protect\independenT}{\perp}}
\def\independenT#1#2{\mathrel{\rlap{$#1#2$}\mkern2mu{#1#2}}}

\begin{document} 
 
\begin{singlespacing}
\maketitle  
\end{singlespacing}
\thispagestyle{empty}

\begin{abstract}  
\noindent 
This paper proposes a method for estimating the effect of a policy intervention on an outcome over time. We train recurrent neural networks (RNNs) on the history of control unit outcomes to learn a useful representation for predicting future outcomes. The learned representation of control units is then applied to the treated units for predicting counterfactual outcomes. RNNs are specifically structured to exploit temporal dependencies in panel data, and are able to learn negative and nonlinear interactions between control unit outcomes. We apply the method to the problem of estimating the long-run impact of U.S. homestead policy on public school spending.

\begin{singlespace} 
	\emph{Keywords:} Counterfactual Prediction; Panel Data; Political Economy; Recurrent Neural Networks; Synthetic Controls
\end{singlespace}
\end{abstract}

\pagebreak
\pagenumbering{arabic}

\section{Introduction} 
\noindent 
An important problem in the social sciences is estimating the effect of a binary treatment on a continuous outcome in a panel data setting. Two prevalent methods for causal inference with panel data are difference-in-differences (DID) and the synthetic control method (SCM). DID uses time-varying panel data to control for time-invariant unobserved confounding, and identifies causal effects by contrasting the change in average outcomes pre- and post-treatment, between treated and control units \citep[e.g.,][]{ashenfelter1978estimating}. DID assumes no time-varying unobserved confounding that affects both treatment and outcomes, which is highly restrictive and cannot be empirically tested. Moreover, the linear DID estimator assumes i.i.d. errors, which ignores the temporal aspect of the data and understates standard errors for estimated treatment effects when the regression errors are serially correlated, which can arise when the time-series lengths are not sufficiently long to reliably estimate the data generating process \citep{bertrand2004much}. 

The SCM is a popular method that constructs a convex combination of control units that are similar to a single treated unit in terms of pretreatment outcomes or covariates \citep{abadie2003economic, abadie2010synthetic, abadie2015comparative}. The SCM estimator assumes there is a stable convex combination of the control units that absorbs all time-varying unobserved confounding and may be biased even if treatment is only correlated with time-invariant unobserved confounding, which is equivalent to the DID identification assumption \citep{ferman2016revisiting}. The SCM can be generalized to settings with staggered treatment adoption, where the time of initial treatment varies across multiple treated units \citep{dube2015pooling,benmichael2019synthetic}, and to include features of DID estimation \citep{2018arXiv181104170B,arkhangelsky2019synthetic} or Bayesian estimation \citep{brodersen2015inferring,pang2020bayesian}. 

We propose a method based on recurrent neural networks (RNNs), a class of neural networks that take advantage of the sequential nature of temporal data by sharing model parameters across multiple time periods \citep{el1995,graves2012}. RNNs have been shown to outperform various linear models on time-series prediction tasks \citep{cinar2017position}. Unlike the SCM, RNNs are able to learn negative and nonlinear interactions between control unit outcomes, and do not assume a specific activation function when learning representations of the control unit outcomes. RNNs are end-to-end trainable, whereas each component of the Bayesian structural time-series model proposed by \citet{brodersen2015inferring} must be assembled and estimated independently. RNNs are capable of sharing learned model weights for predicting multiple treated units, and can thus generate more precise predictions in settings where treated units share similar data-generating processes. 

We train RNNs on the control units outcomes data to learn a useful latent representation of outcomes in previous periods for predicting future outcomes. We weight the RNNs loss function by the propensity scores modeled in terms of pretreatment covariates to ensure the weighted distribution of the observed confounders are balanced between treated and control units. The learned representation of control unit outcomes is then applied to the outcomes data of the treated units for predicting counterfactual outcomes. The causal effect of treatment on the treated units is estimated by contrasting the counterfactual predictions to the observed outcomes of the treated. 

The RNN-based method is related to lagged regression models, which regress post-treatment outcomes on pretreatment outcomes and covariates for control units and then use the model weights to predict the counterfactual outcome for treated units \citep[e.g.,][]{athey2016approximate,belloni2017program,carvalho2018arco}. A closely related approach are linear factor models, which decompose the pretreatment outcomes of control units into matrices of latent factors (i.e., time-varying coefficients) and factor loadings (i.e., unit-specific intercepts) and predict counterfactual treated unit outcomes based on the estimated factors and loadings \citep[e.g.,][]{xu2017generalized,athey2017matrix,amjad2018robust}. These models typically use regularization or matrix factorization to reduce the dimensionality of the predictor set and thereby improve generalizability when applying the model fit on control units to treated units. These methods all assume unconfoundedness conditional on previous outcomes for control units.

The proposed method is also related to doubly-robust estimators that combine both a propensity score model and an outcome model, which are consistent if either model is properly specified \citep{bang2005doubly,10.1111/ectj.12097}. Several studies independent of this work propose using neural networks for counterfactual prediction in non-panel observational data settings. For example, \citet{farrell2018deep} provide inference results for semiparametric estimation of causal effects using multilayer perceptrons, while \citet{pmlr-v70-hartford17a} and \citet{NIPS2019_8615} integrate deep neural networks into an instrumental variables framework. 

In the section immediately below, we state the problem of counterfactual prediction within the potential outcomes framework; Section \ref{RNNs-section} introduces the approach of using RNNs for counterfactual prediction; Section \ref{placebo} presents the results of placebo tests; Section \ref{schooling-app} applies the method to the problem of estimating the long-run impact of U.S. homestead policy on state government investment in public schooling; Section \ref{conclusion} concludes and offers potential avenues for future research. 

\section{Potential outcomes framework} \label{counterfactual-prediction}

We explore a panel data setting where we observe a real-valued continuous outcome $Y_{it}$ for each $i = 1, \ldots, N$ units and in each $t = 1, \ldots, T$ time periods, and where a subset of units is exposed to a binary treatment $W_{it} \in \left\{0,1\right\}$ following an initial treatment period $T_{0}$. We also observe time-invariant pre-treatment covariates, $V_{ip}$, where $p$ denotes the number of predictors. 

We follow the Neyman-Rubin potential outcomes framework \citep{neyman1923,rubin1974estimating,rubin1990}, where there exists a pair of potential outcomes, $Y_{it}(1)$ and $Y_{it}(0)$, corresponding to the response to treatment and control, respectively. The potential outcomes framework implicitly assumes treatment is well-defined to ensure that each unit has the same number of potential outcomes. It also excludes interference between units, which would undermine the framework by creating more than two potential outcomes per unit, depending on the treatment status of other units. We only observe one of the two potential outcomes for each $it$ value, while the other outcome is counterfactual. The observed outcomes are:
\begin{equation}Y_{it}=\left\{\begin{array}{ll}
		Y_{it}(0) & \text { if } W_{i}=0 \text { or } t < T_{0} \\
		Y_{it}(1) & \text { if } W_{i}=1 \text { and } t \geq T_{0}.
	\end{array}\right.
\end{equation}

Define $\tau_{it} = Y_{it}(1) - Y_{it}(0)$ as the individual treatment effect. The causal estimand of interest is the Average Treatment Effect on the Treated (ATT):
\begin{equation}
\tau_t^{\text{ATT}} = \E\left[\tau_{it} | W_{it} = 1\right], \text{ for }  t \in \left\{T_{0}, \ldots, T\right\}. \label{eq:att-estimand}
\end{equation}
In the empirical application, we focus on estimating the ATT averaged over the post-treatment period:
\begin{equation}
\tau^{\textup{ATT}} =\sum_{t=T_{0}}^{T}\tau_{t}^{\textup{ATT}}/(T-T_{0}+1). \label{eq:att-avg}
\end{equation}
To estimate $\tau_t^{\textrm{ATT}}$, we predict $Y_{it}(0)$ for all $it$ values with $W_{it} =1$; i.e., the counterfactual outcome of the treated units had they not been exposed to treatment. The counterfactual predictions are subsequently plugged into the estimator:
\begin{equation}
\hat{\tau}_t^{\textrm{ATT}} =\frac{\sum_{it} W_{it} \left(Y_{it} (1) - \hat{Y}_{it}(0)\right)}{\sum_{it} W_{it}}. \label{eq:att}
\end{equation}
The causal estimand $\tau_t^{\text{ATT}}$ is identified by assuming that treatment and potential outcomes under control are unconfounded conditional on the pre-treatment outcomes and covariates. 
\begin{assumption}{1}{}\label{unconfoundedness}
	Conditional unconfoundedness:
	\begin{equation*}
	W_{it} \independent Y_{it}(0)  \mid Y_{i,1}, \ldots, Y_{i,T-1}, V_{ip}.
	\end{equation*}
\end{assumption}
\noindent
Assumption \eqref{unconfoundedness} ensures that treatment assignment affects potential outcomes under control only through covariates and the history of observed outcomes. The idea is that a hypothetical conditional randomization is taking place but, differently from observational studies under the usual strong ignorability assumption \citep[Ch.~12]{imbens2015causal}, the conditioning set includes the outcome history. 
\begin{assumption}{2}{}\label{overlap}
	Overlap:
	\begin{equation*}
	 0 < e_{it} < 1, \qquad \mathrm{where} \qquad e_{it} = \mbox{Pr} (W_{it} =1 | Y_{i,1}, \ldots, Y_{i,T_0-1}, V_{ip}).
	\end{equation*}
\end{assumption}
\noindent
Assumption \eqref{overlap} is needed to summarize the treatment assignment mechanism by the propensity score, $e_{it}$. As described in Section \ref{training}, we use the estimated propensity score to weight the RNNs loss function to correct for imbalances in the distributions of the conditioning set between the treated and control units. 

\section{RNNs for counterfactual prediction} \label{RNNs-section}
\noindent
RNNs operate on $n$ inputs $X=\left(\mathbf{x}^{1}, \mathbf{x}^{2}, \cdots, \mathbf{x}^{n}\right)^{\top}=\left(\mathbf{x}_{1}, \mathbf{x}_{2}, \cdots, \mathbf{x}_{\textrm{T}_x}\right) \in \mathbb{R}^{n \times \textrm{T}_x}$, where $\textrm{T}_x$ is the input sequence length. The task is to predict outputs $Y=\left(\mathbf{y}_{1}, \mathbf{y}_{2}, \cdots, \mathbf{y}_{\textrm{T}_y}\right) \in \mathbb{R}^{n \times \textrm{T}_y}$, where output length $\textrm{T}_y$ can differ from $\textrm{T}_x$, and $\textrm{T}_x + \textrm{T}_y = \textrm{T}$. Given $X$, we aim to learn a nonlinear mapping $\mathcal{F}(\cdot)$ to predict the next values of the output sequence, $\hat{y}_{t+1}=\mathcal{F}\left(X\right)$. RNNs capture nonlinear correlations of the historical values of $Y_{it}$ for $t = 1,\ldots, T-1$ to the future values of $Y_{it}$. The parameter sharing used in RNNs assumes the same learned model parameters are shared for all $t$; that is, the estimated nonlinear correlations are stationary. 

\subsection{Training process}\label{training}

At each $t$, the RNNs input $\boldsymbol{x}_{t}$ and pass it to a fixed-length vector $\boldsymbol{h}_{t}$ called the hidden state, which  stores information from the history of inputs up to $\boldsymbol{x}_t$ and the previous hidden state, $\boldsymbol{h}_{t-1}$. Starting with an initial hidden state $\boldsymbol{h}_{0}$, $\boldsymbol{h}_{t}$ is updated from $t=1, \ldots \textrm{T}_x$ according to the forward propagation equations:
\begin{align}
\boldsymbol{a}_{t} &= b + Q\boldsymbol{h}_{t-1} + R\boldsymbol{x}_{t}  \label{eq:activations}\\
\boldsymbol{h}_{t} &= f \left(\boldsymbol{a}_{t}\right) \label{eq:hidden} \\
\boldsymbol{\hat{y}}_{t} &= d + U\boldsymbol{h}_{t}, \label{eq:output}
\end{align} 
where $Q$, $R$, and $U$ are weight matrices, and $b$ and $d$ are constants. The constants are initialized at zero and $\boldsymbol{h}_{0}$ is initialized by drawing values from a uniform distribution \citep{glorot2010}. The hidden state activation function in Eq. \eqref{eq:hidden} is a nonlinear function such as the hyperbolic tangent
(tanh), which is a shifted and scaled version of the logistic function that is commonly used with RNNs because the gradient computation is cheaper compared to the logistic function \citep{sochercs224d}. 

In Eqs. \eqref{eq:activations} and \eqref{eq:hidden}, the activation function computes the value of $\boldsymbol{h}_{t}$ using information from the previous hidden state $\boldsymbol{h}_{t-1}$ and the current input $\boldsymbol{x}_{t}$. The parameters used to compute $\boldsymbol{h}_{t}$ are shared for each value of $t$. In Eq. \eqref{eq:output}, the RNNs read information from $\boldsymbol{h}_{t}$ to output a sequence of predicted values $\boldsymbol{\hat{y}}^{(t)}$. The process of forward propagation culminates in producing a loss that compares $\boldsymbol{\hat{y}}^{(t)}$ and $\boldsymbol{y}^{(t)}$. Gradients for Eq. \eqref{eq:activations} and Eq. \eqref{wmse} are computed by the back-propagation through time algorithm \citep[p.~384]{goodfellow2016deep}, which are subsequently used for gradient descent to estimate the network parameters. 

We weight the RNNs objective function to minimize the MSE weighted by the propensity score, which we estimate by multiresponse lasso regression in order to share model parameters across multiple time periods and to shrink the coefficients of (all but one) correlated covariates towards zero \citep{tibshirani2012strong, simon2013blockwise}. In order avoid extreme propensity weights, we employ overlap weighting so that observed values under treatment receive a weight of $1-\hat{e}_{it}$ and observed values under control receive a weight equal to $\hat{e}_{it}$ \citep{li2018balancing}. The propensity-score weighted MSE is:
\begin{equation} \label{wmse}
\text{L} = \frac{1}{T_y} \sum^{T_y}_{t=1}  W_{it} \,(1-\mathbf{\hat{\text{e}}}_{it}) + (1-W_{it})\mathbf{\hat{\text{e}}}_{it} \, \left(\boldsymbol{\hat{y}}_{t} - \boldsymbol{y}_{t}\right)^2 +  \lambda \mathbf{u}_t^2, 
\end{equation}
where the right-hand-side term is the ridge penalty on the learned weights, $U = \left(\mathbf{u}_1, \mathbf{u}_2, \ldots, \mathbf{u}_{T_y}\right)$ from Eq.~\eqref{eq:output}, and $\lambda > 0$ controls the regularization strength.

We train RNNs on control outcomes using a sliding window with step size of one. Each sliding window contains 10 time periods as input, and aims to predict following time period. Data are fed into the networks in batches of size 32 and the networks are trained for 500 epochs with mini-batch gradient descent on Eq.~\eqref{wmse}. The last 20\% of the training set is reserved for model validation. In order to prevent over-fitting, the training process stops early when there ceases to be improvement on the validation set loss within 25 epochs, and in the event of early stopping, the model weights associated with the lowest validation set loss are restored. In addition to applying a ridge penalty to Eq.~\eqref{wmse}, we regularize the networks by applying dropout to both the hidden units and recurrent connections \citep{gal2015theoretically}. 

\subsection{Identification} \label{identification}

After training, we can impute the missing potential outcomes under control $\widehat{Y_{it}}(0)$ with the forward propagation equations in \eqref{eq:activations}-\eqref{eq:output}. We train on the control units to predict the missing outcome $\widehat{{Y}_{it}}(0)$ for those $it$ values with $W_{it}=1$. We then estimate $\tau_{t}^{\textup{ATT}}$ \eqref{eq:att} by contrasting $\widehat{Y_{it}}(0)$ with $Y_{it}(1)$. Under Assumptions \eqref{unconfoundedness} and \eqref{overlap}, $\tau_{t}^{\textup{ATT}}$ can be identified.

\subsection{Network architecture} \label{encoder-decoder}

We employ encoder-decoder networks, which are the standard for neural machine translation \citep{cho2014learning,bahdanau2014neural,vinyals2014grammar} and are also widely used for predictive tasks, including speech recognition \citep[e.g.,][]{chorowski2015attention} and time-series forecasting \citep[e.g.,][]{zhu2017deep}. Encoder-decoder networks consist of an encoder and decoder RNN, both taking the form of long short-term memory (LSTM) networks. LSTMs are designed to resolve problems such as vanishing and exploding gradients, which prevent the networks from learning long-term dependencies in the data and tends to occur when the dimension of the hidden states is too small to summarize long input sequences \citep{pascanu2013difficulty,bahdanau2014neural}.

The encoder RNN reads in $X$ sequentially and the hidden state of the network updates according to Eq. \eqref{eq:hidden}. The final hidden state of the encoder is a fixed-size context vector $\boldsymbol{c}$ that summarizes the input sequence, which is copied over to the decoder RNN. Thus, the hidden state of the decoder is updated recursively by
\begin{equation}
\boldsymbol{h}_{t} = f \left( \boldsymbol{h}_{t-1}, \boldsymbol{y}_{t-1}, \boldsymbol{c}; \, \theta \right), \label{eq:hidden-decoder}
\end{equation} 
and the conditional probability of the next element of the sequence is 
\begin{equation}
\Pr(\boldsymbol{y}_{t} \mid \boldsymbol{y}_{t}, \ldots, \boldsymbol{y}_{t-1}, \boldsymbol{c}) = f \left( \boldsymbol{h}_{t-1}, \boldsymbol{y}_{t-1}, \boldsymbol{c}; \, \theta \right).
\end{equation} 
As in Eq.~\eqref{eq:output}, the decoder uses $\boldsymbol{h}_{t}$ to predict $\boldsymbol{y}_{t}$ at each $t$. 

In our experiments and empirical application, the encoder-decoder networks consist of a two-layer LSTM encoder and single-layer Gated Recurrent Unit (GRU) \citep{chung2014} decoder, each with 128 hidden units and tanh activation, stacked on a fully-connected output layer with no activation function, i.e., $f(x) = x$. We compare the encoder-decoder networks with a baseline LSTM consisting of a single-layer LSTM stacked on a fully-connected output layer. The baseline LSTM has fewer network parameters than the encoder-decoder networks, and is expected to be more suitable for smaller-dimensional datasets. 

\section{Placebo test experiments} \label{placebo} 

We conduct a series of placebo test experiments on data without real interventions in order to evaluate the ability of the RNN-based estimator to recover a null average treatment effect, i.e., $\tau_t^{\text{ATT}} = 0$. The rationale for placebo test experiments is that we know the ground-truth and Assumption~\eqref{unconfoundedness} is satisfied in expectation because placebo treatment is assigned randomly. For each trial run, we randomly select $N/2$ placebo treated units and predict their outcomes for periods following a given placebo initial time period under a staggered treatment adoption setting. 

We benchmark the performance of the encoder-decoder networks described in Section \ref{encoder-decoder} and a baseline single-layer LSTM against several estimators in terms of the root mean squared error (RMSE), comparing the actual and predicted values. A full description of the benchmark estimators we consider are provided in the supporting material (SM), Section \ref{benchmark-estimators}. 
\begin{singlespace}
	\begin{description}
		{\setlength\itemindent{1mm}
			\item[(a) DID] Difference-in-differences regression of outcomes on treatment and unit and time fixed effects \citep{athey2018design};
			\item[(b) MC-NNM] Nuclear norm regularized matrix completion estimator \citep{athey2017matrix};
			\item[(c) SCM] Generalized SCM with the restriction of nonnegative weights and zero intercept of the original SCM, and weights estimated by gradient descent \citep{doudchenko2016balancing};
			\item[(d) SCM-L1] Generalized SCM with an intercept and without weights restrictions, and weights estimated by lasso linear regression \citep{doudchenko2016balancing};
			\item[(e) VAR] Stationary vector autoregression, with weights estimated by lasso linear regression \citep{kock2015oracle}.
		}
	\end{description}
\end{singlespace}

The comparison between the RNN-based estimators and the VAR is particularly important because the latter is capable of modeling a linear dependency component, which is present in most real-world datasets, while the former is suitable for modeling a nonlinear dependency component that potentially spans across multiple time periods \citep{DBLP:journals/corr/abs-1709-03159}. However, RNNs typically require a large amount of training data to effectively capture nonlinear and potentially long-term dependencies. 

To facilitate the comparison, we run placebo test experiments on the dataset underlying our empirical application along with three high-dimensional datasets. The education spending dataset, described in Section \ref{educ-data}, consists of historical data on the per-capita education spending of 36 U.S. state governments over $\text{T} = 203$ years. We remove the treated units from the dataset, leaving $N=18$ control states. The Gaussian processes data are smooth signals generated using radial basis function to specify a Gaussian process with zero-valued mean function. The sine waves data are generated nonlinear variations with frequencies in $[1.0, 5.0]$, amplitudes in $[0.1, 0.9]$, and random phases between $[-\pi, \pi]$. The Gaussian processes and sine waves datasets each consist of $N=4,956$ samples with sequence length of $T = 500$. Lastly, the stock prices dataset consists of stock market returns for $\text{N} = 2,453$ stocks over $\text{T} = 3,082$ days. The stock market is dynamic, non-stationary and complex in nature, and predicting stock market returns is a challenging task due to its unpredictable and nonlinear nature. 

\begin{figure}[htb]
	\centering
	\includegraphics[width=\textwidth]{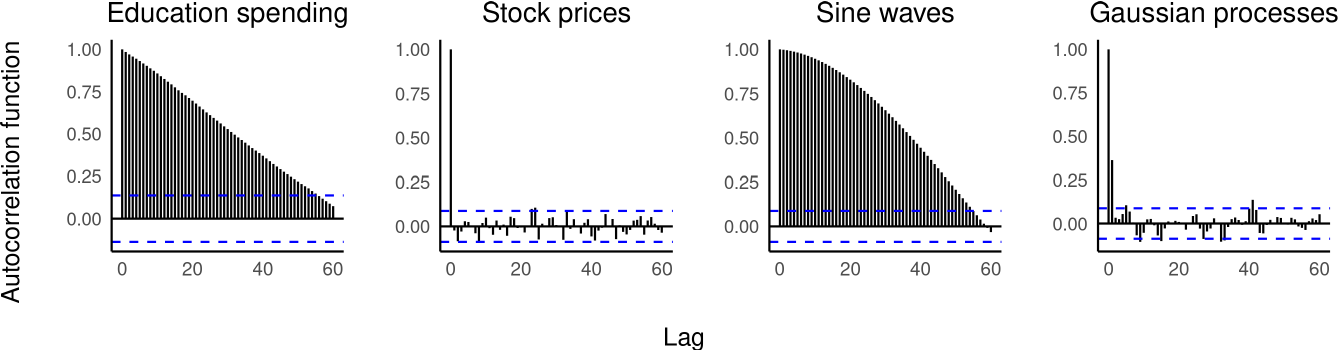}
	\caption{Autocorrelation function of placebo test datasets. Dashed horizontal lines represent 95\% confidence bands.}
	\label{fig:acf_placebo}
\end{figure}

In Figure \ref{fig:acf_placebo}, we calculate the autocorrelation function $\rho(k)=\textup{Cov}(Y_{t},Y_{t+k})/var(Y_{t})$ for the observations in different placebo test datasets and a given lag $k$. The correlation decays quickly in the stock prices and Gaussian processes datasets, which suggests that the outcomes depend mostly on the observations in the short run. On the other hand, the correlation in the education spending and sine waves datasets remain prominent for the order lagged before 55 and 54, respectively, indicating that the dependency along the time dimension is longer term and of a more complicated structure. The RNN-based estimators and VAR are expected to hold an advantage over the other estimators in capturing the temporal information or dependency in the long-run.

\begin{figure}[htbp]
	\centering
	\begin{subfigure}{0.5\textwidth}
		\includegraphics[width=\textwidth]{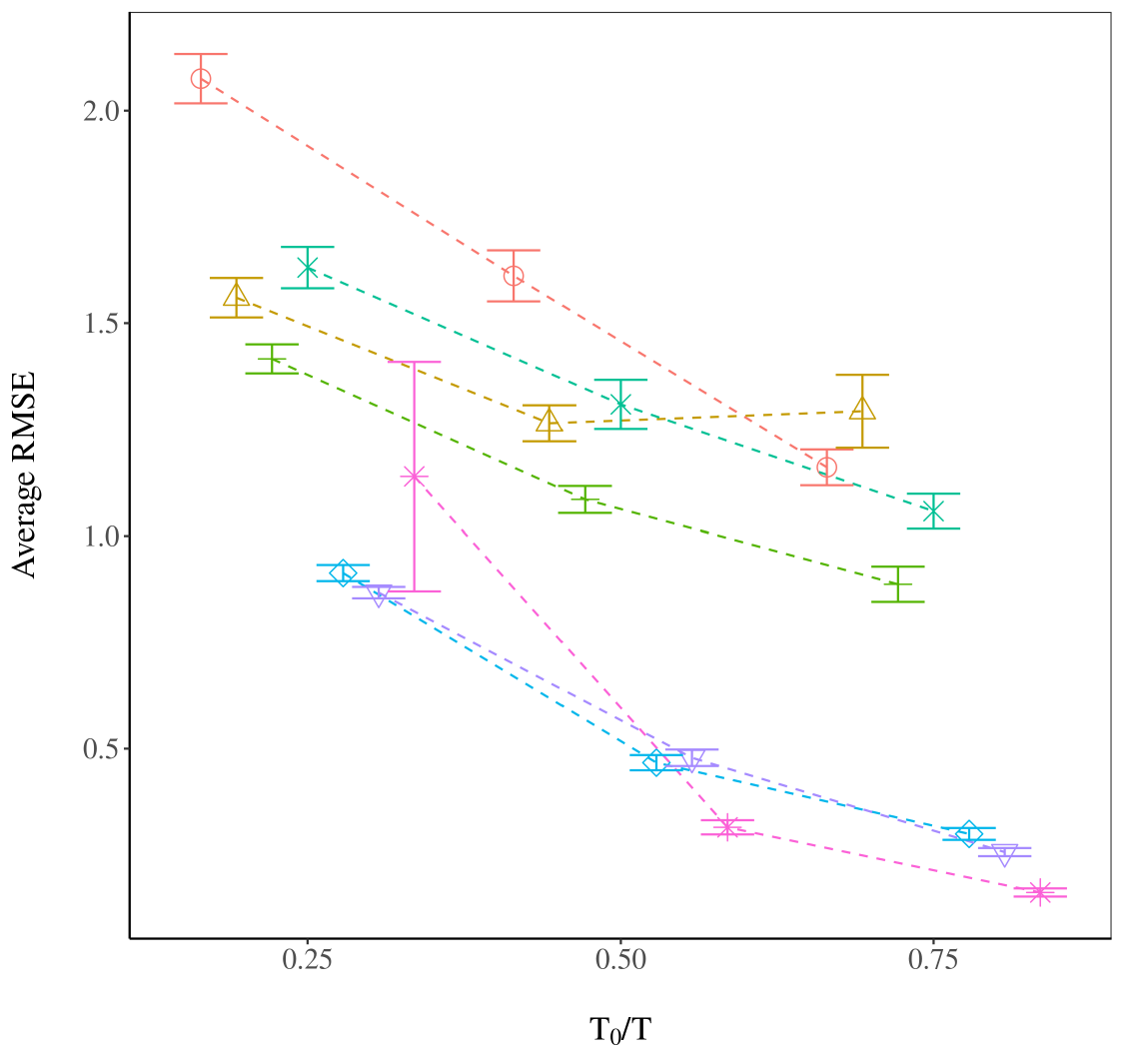}
		\caption{Education spending: $(N \times T) = (18 \times 203)\dagger$.}
	\end{subfigure}%
	~ 
	\begin{subfigure}{0.5\textwidth}
		\includegraphics[width=\textwidth]{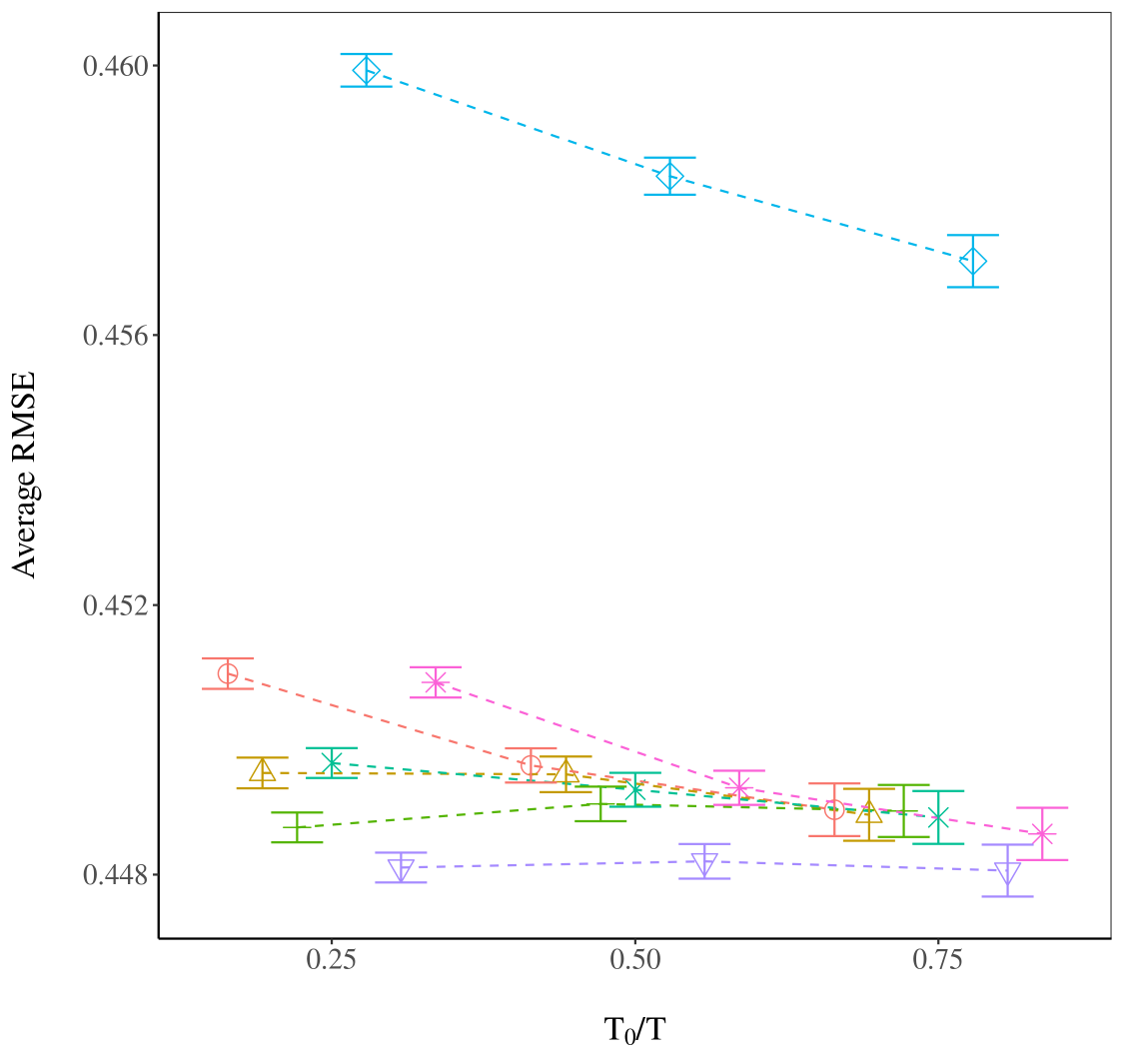}
		\caption{Gaussian processes: $(N \times T) = (1,000 \times 500)\ddagger$.}
	\end{subfigure}%
	\hfill
	\caption{Placebo tests under staggered treatment adoption. Vertical lines represent $\pm$ 1.96 times the standard error of the average RMSE across 100 runs. $\dagger$: Subset from $(N \times T) = (38 \times 203)$; $\ddagger$: sub-sampled from $(N \times T) = (4,956 \times 500)$. Method: 
		{\protect\tikz \protect\draw[color={rgb:red,248;green,118;yellow,109}] (0,0) -- plot[mark=o, mark options={scale=3}] (0.25,0) -- (0.5,0);}, DID;
		{\protect\tikz \protect\draw[color={rgb:red,196;green,154;blue,0}] (0,0) -- plot[mark=triangle*, mark options={scale=3,fill=white}] (0.25,0) -- (0.5,0);}, Encoder-decoder (ours); 
		{\protect\tikz \protect\draw[color={rgb:red,83;green,180;blue,0}] (0,0) -- plot[mark=+, mark options={scale=3}] (0.25,0) -- (0.5,0);}, LSTM (ours);
		{\protect\tikz \protect\draw[color={rgb:red,0;green,192;blue,148}] (0,0) -- plot[mark=x, mark options={scale=3, rotate=180}] (0.25,0) -- (0.5,0);}, MC-NNM;
		{\protect\tikz \protect\draw[color={rgb:red,0;green,182;blue,235}] (0,0) -- plot[mark=diamond, mark options={scale=3, rotate=180}] (0.25,0) -- (0.5,0);}, SCM;
		{\protect\tikz \protect\draw[color={rgb:red,165;green,138;blue,255}] (0,0) -- plot[mark=triangle, mark options={scale=3, rotate=180}] (0.25,0) -- (0.5,0);}, SCM-L1;
		{\protect\tikz \protect\draw[color={rgb:red,251;green,97;blue,215}] (0,0) -- plot[mark=asterisk, mark options={scale=3}] (0.25,0) -- (0.5,0);}, VAR.\label{educ-sim}} 
\end{figure}

Figure \ref{educ-sim} reports average RMSE for each estimator the education spending and Gaussian processes datasets, varying the placebo initial treatment time under randomly assigned treatment in a staggered treatment adoption setting. The horizontal axis is the ratio of the initial placebo treatment time to the number of periods in the placebo data, so higher values represent more training data, and estimates are jittered across the horizontal axis to avoid overlap. When trained on the education spending data, the average RMSE for the RNN-based estimators generally decreases as the amount of training data increases, reflecting the need for sufficient time periods. The LSTM and encoder-decoder networks outperform DID and matrix completion on the education spending data, and perform comparatively to the SCM estimators and VAR when trained on the higher-dimensional Gaussian processes data.

\begin{figure}[htbp]
	\centering
	\includegraphics[width=\textwidth]{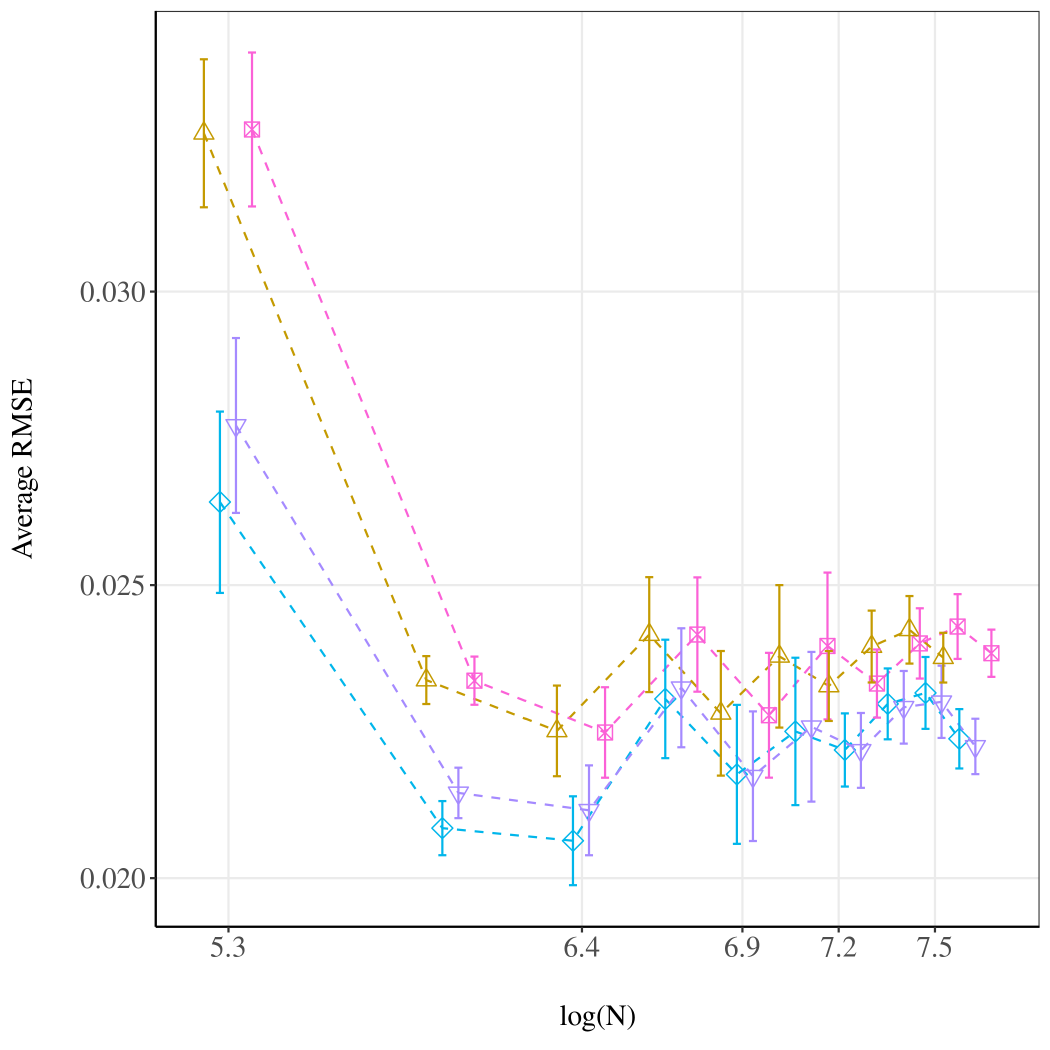}
	\hfill
	\caption{Stock prices data: placebo tests under staggered treatment adoption, with varying dimensions and keeping $N \times T=400,000$. Vertical lines represent $\pm$ 1.96 times the standard error of the average RMSE across 10 runs. \emph{Method:}
		{\protect\tikz \protect\draw[color={rgb:red,205;green,150;blue,0}] (0,0) -- plot[mark=triangle*, mark options={scale=3,fill=white}] (0.25,0) -- (0.5,0);}, Encoder-decoder (ours); 
	{\protect\tikz \protect\draw[color={rgb:red,0;green,182;blue,235}] (0,0) -- plot[mark=diamond, mark options={scale=3, rotate=180}] (0.25,0) -- (0.5,0);}, SCM;
	{\protect\tikz \protect\draw[color={rgb:red,165;green,138;blue,255}] (0,0) -- plot[mark=triangle, mark options={scale=3, rotate=180}] (0.25,0) -- (0.5,0);}, SCM-L1;
	{\protect\tikz \protect\draw[color={rgb:red,251;green,97;blue,215}] (0,0) -- plot[mark=asterisk, mark options={scale=3}] (0.25,0) -- (0.5,0);}, VAR. \label{stock-sim}} 
\end{figure}

In Figure \ref{stock-sim}, we create 10 different sub-samples by selecting the first $T$ daily returns of $N$ randomly selected stocks, keeping the overall data dimension fixed at $N \times T = 400,000$, focusing on encoder-decoder networks, SCM estimators, and VAR. The sub-sampled matrices range from very thin, $N \times T = (200 \times 2,000)$, to very fat, $N \times T = (2,000 \times 200)$. In each case, $N/2$ units are randomly selected for placebo treatment starting at an initial placebo time period of $T/2$. The average RMSE is the highest for the encoder-decoder networks when the data are very thin, which reflects the benefit of training on high-dimensional data. All estimators perform comparatively across the remaining sub-samples.

Table \ref{tab:stag-rmse-table} reports the average RMSE for the estimators with the placebo initial treatment period set to $T/2$. For the stock prices data, the first $T=500$ daily returns of $N=1,000$ stocks are randomly sub-sampled from the larger dataset. The baseline LSTM outperforms the encoder-decoder networks in the education spending and sine waves datasets, underscoring how the higher complexity of the encoder-decoder networks can hinder performance on datasets with smaller dimensions or those with simple signals, such as with the sine waves data. The VAR is capable of capturing the linear interdependencies among multiple time series and achieves the lowest average RMSE on the education spending dataset, whereas
the SCM estimators perform better on the other datasets. The RNN-based estimators perform comparably to the SCM and VAR estimators in the Gaussian processes, sine waves, and stock prices datsets, and outperform both DID and matrix completion estimators in the education spending and sine waves datasets.

\begin{table}[htbp]
	\centering
	\begin{adjustbox}{max width=\textwidth}
		\begin{threeparttable}	\footnotesize
			\captionsetup{font=normalsize}
			\caption{Average RMSE on test set under staggered treatment adoption.\label{tab:stag-rmse-table}}
			\begin{tabular}{@{}lccccc@{}}
		\toprule
		\multicolumn{1}{l}{} &
		\multicolumn{1}{c}{\begin{tabular}[c]{@{}c@{}} Education spending\\ $(18 \times 203)\dagger$\end{tabular}} &
		\multicolumn{1}{c}{\begin{tabular}[c]{@{}c@{}} Gaussian processes \\$(1,000 \times 500)\ddagger$ \end{tabular}} &
		\multicolumn{1}{c}{\begin{tabular}[c]{@{}c@{}} Sine waves\\$(1,000 \times 500)\ddagger$ \end{tabular}} &
		\multicolumn{1}{c}{\begin{tabular}[c]{@{}c@{}} Stock prices\\$(1,000 \times 500)\mathsection$ \end{tabular}} \\ \midrule
		\hline \\
		DID 			& 1.611 $\pm$ 0.030 & 0.449 $\pm$ $10^{-4}$	& 0.467 $\pm$ 0.001	& 0.023 $\pm$ $2 \times 10^{-4}$	\\
		Encoder-decoder (ours) & 1.264 $\pm$ 0.021	& 0.449 $\pm$ $10^{-4}$	& 0.405 $\pm$ 0.001	& 0.023 $\pm$ $10^{-4}$	\\
		LSTM (ours) 			& 1.085 $\pm$ 0.016 & 0.449 $\pm$ $10^{-4}$	& 0.398 $\pm$ 0.001	& 0.023 $\pm$ $10^{-4}$	\\
		MC-NNM     		& 1.308 $\pm$ 0.029 & 0.449 $\pm$ $10^{-4}$	& 0.457 $\pm$ 0.001	& 0.023 $\pm$ $2 \times 10^{-4}$ \\
		SCM     & 0.467 $\pm$ 0.009			& 0.458 $\pm$ $10^{-4}$	& 0.349 $\pm$ 0.001	& $\mathbf{0.022 \pm 2 \times 10^{-4}}$ \\
		SCM-L1  & 0.478 $\pm$ 0.009 		& $\mathbf{0.448 \pm 10^{-4}}$	& $\mathbf{0.285 \pm 0.001}$	& $0.023 \pm 2 \times 10^{-4}$ \\
		VAR     & $\mathbf{0.314 \pm 0.008}$  		& 0.449 $\pm$ $10^{-4}$	& 0.326 $\pm$ 0.002	& 0.024 $\pm$ $10^{-4}$ \\ \bottomrule
	\end{tabular}
			\begin{tablenotes}[flushleft]
				\small
				\item \emph{Notes:} Average RMSE $\pm$ 1.96 times the standard error across 100 runs, with $N/2$ treated units and $T/2$ treated periods. $\dagger$: Subset from $(N \times T) = (38 \times 203)$; $\ddagger$: sub-sampled from $(N \times T) = (4,956 \times 500)$; $\mathsection$: sub-sampled from $(N \times T) = (2,453 \times 3,082)$. Bold indicates lowest average RMSE.
			\end{tablenotes}
		\end{threeparttable}
	\end{adjustbox}
\end{table}

\section{Homestead policy and public schooling in the U.S.} \label{schooling-app}

In the empirical application, we are interested in estimating the impact of mid-19th century homestead policy on the development of state government public education spending in the U.S. Social scientists have long viewed the rapid development of public schooling in the U.S. as a nation-building policy \citep[e.g,][]{meyer1979public,alesina2013nation,bandiera2018nation}. According to this view, states across the U.S. adopted compulsory primary education to homogenize the population during the Age of Mass Migration, when tens of millions of foreign migrants arrived to the country between 1850 and 1914. 

\citet{engerman2005evolution} propose an alternative explanation for the rise of public schooling: state governments on the western frontier expanded investments in public education to attract eastern migrants following the passage of the Homestead Act (HSA) of 1862. The HSA opened for settlement hundreds of millions of acres of frontier land, and any adult citizen could apply for a homestead grant of 160 acres of land, provided that they live and make improvements on the land for five years. According to this view, the sparse population on the frontier meant that state governments competed with each other to attract migrants in order to lower local labor costs and to increase land values and tax revenues. State governments in public land states --- i.e., states crafted from the public domain and open to homesteading --- offered migrants broad access to cheap land and property rights, unrestricted voting rights, and access to public schooling. State land states, which include the original 13 states, Kentucky, Maine, Tennessee, Texas, Vermont, and West Virginia, were not open to homesteading because the state government had primary authority to distribute public land \citep{murtazashvili2013political}. 

Another alternative view is that the HSA led to larger investments in public schooling by reducing the degree of land inequality on the frontier as a consequence of fixing land grants to 160 acres. Political economy frameworks \citep[e.g.,][]{acemoglu2008persistence, besley2009origins} emphasize that greater economic power of the ruling class reduces public investments. In the model of \citet{galor2009inequality}, wealthy landowners block education reforms because public schooling favors industrial labor productivity and decreases the value in farm rents. Inequality in this context can be thought of as a proxy for the amount of \emph{de facto} political influence elites have to block education reforms.

\subsection{Data} \label{educ-data}

We draw data on state government education spending from the records of 48 state governments during the period of 1789 to 1932 \citep{sylla1993sources}, 16 state governments during the period of 1933 to
1937 \citep{sylla1995sourcesa,sylla1995sourcesb}, and U.S. Census special reports for the years 1902, 1913, 1932, 1942, 1944, 1946, 1948, and 1950 to 2008, covering 48 states \citep{haines2010,census2010}. Removing states and years with zero or near-zero variance results in a dataset consisting of $T=203$ observations for $N=38$ U.S. states, half of which are treated. We inflation-adjust the education spending data according to the U.S. Consumer Price Index and scale by the total free population in the decennial census. We impute the 34.1\% of values in the dataset that are missing by Last Observation Carried Forward (LOCF), which replaces each missing value with the most recent non-missing value prior to it, with remaining missing values carried backward. We visualize the extent of the missing data in Figure \ref{fig:missing-heatmap} and evaluate the sensitivity of the causal estimates to alternative imputation methods in Section \ref{imp-sens}. Lastly, we log-transform the data to alleviate exponential effects. 

The staggered treatment adoption setting is appropriate for this application because $\text{T}_0$ varies across states that were exposed to homesteads following the passage of the HSA. We aggregate approximately 1.46 million records of individual land patents authorized under the HSA to the state level in order to determine how the initial treatment time varies across states \citep{GLO}. Using these records, we determine that the earliest homestead patents were filed in 1869 in about half of the public land states, while the remaining public land states had patents filed in subsequent years.

To minimize the discrepancy between the covariate distributions of public land states and state land states, we weight the training loss by propensity scores given per-capita education spending during pre-treatment years. We also include in the conditioning set state-level averages of farm sizes and farm values measured in the 1850 and 1860 censuses \citep{haines2010} and the state-level share of total miles of operational railroad track per square mile \citep{atack2013use}. These pre-treatment covariates control for homesteaders migrating to more productive land and for selection bias arising from differences in access to frontier lands.

While Assumption \eqref{unconfoundedness} cannot be directly tested, the placebo tests on pre-treatment data reported in Section \ref{placebo-pretreatment} provide indirect evidence that unconfoundedness is not violated. The no interference assumption also cannot directly be tested; however, it is likely that state land states were indirectly affected by the out-migration of homesteaders from public land states. Interference in this case would likely cause the estimated treatment effect to be understated.

\subsection{Main estimates} \label{estimates}

\begin{figure}[htbp]
	\centering
	\includegraphics[width=\textwidth]{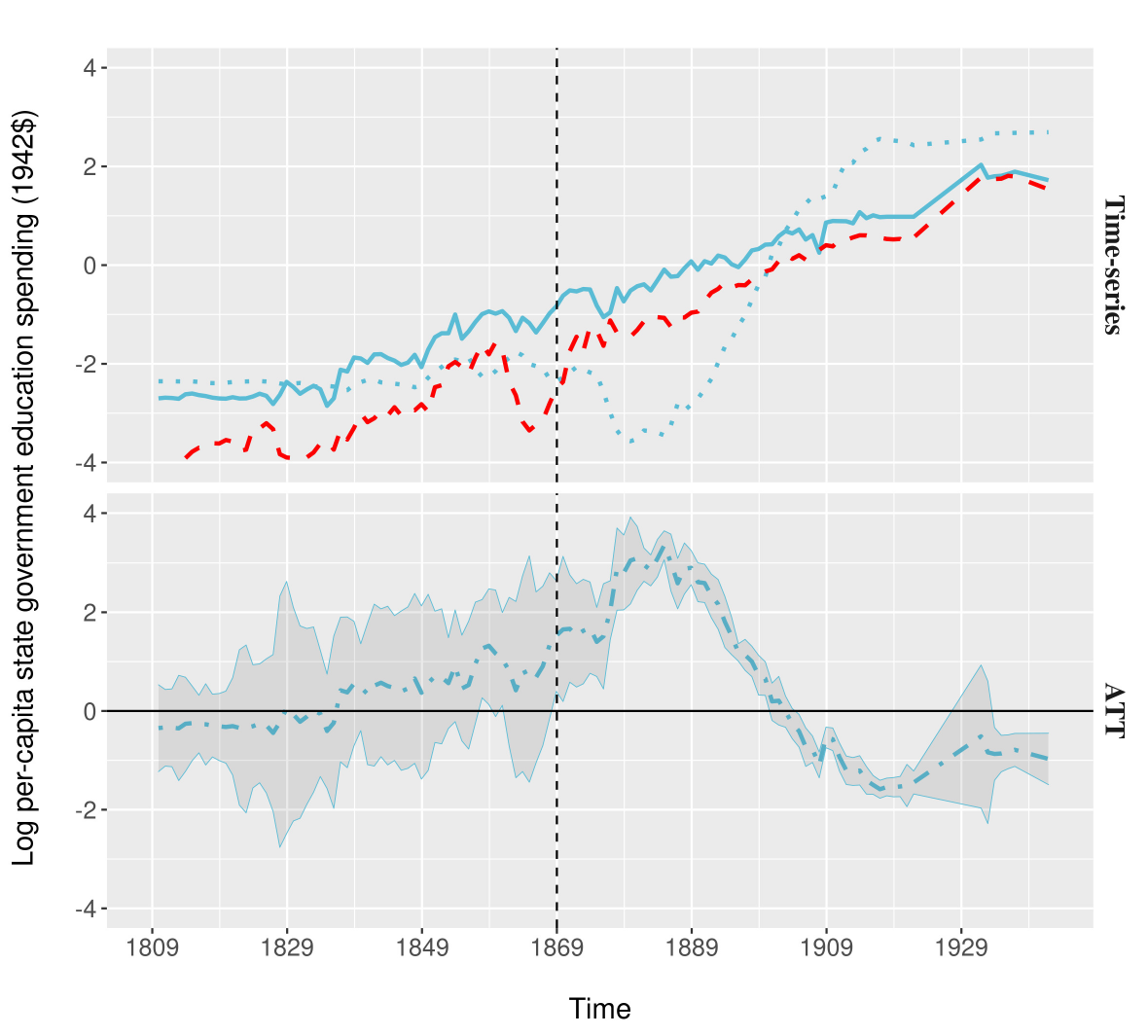}
	\caption{Encoder-decoder estimates of the impact of the HSA on state government education spending, 1809 to 1942, with LOCF imputed missing values. The dashed vertical line represents the first treatment time in 1869. \emph{Key:}		{\color{Darjeeling15}{\sampleline{}}}, observed treated;
		{\color{Darjeeling11}{\sampleline{dashed}}}, observed control;
		{\color{Darjeeling15}{\sampleline{dotted}}}, counterfactual treated;
		{\color{Darjeeling15}{\sampleline{dash pattern=on .7em off .2em on .05em off .2em}}}, $\hat{\tau}_t^{\text{ATT}}$.\label{educ-ed}} 
\end{figure}

We train RNNs on the state land states (i.e., control units) and use the learned weights to predict the counterfactual outcomes of public land states (i.e., treated units). The top panel of Figure \ref{educ-ed} plots the counterfactual predictions of encoder-decoder networks along with the observed outcomes of treated and control units. Prior to first homestead patent in 1869, the predicted outcomes of the public land states closely track their observed outcomes, which indicates that the networks perform well in the prediction task. The bottom panel plots the differences in the observed and predicted outcomes of the public land states, which are bounded by 95\% randomization confidence intervals. We estimate the confidence intervals by constructing a distribution of average placebo effects under the null hypothesis \citep{cavallo2013catastrophic,hahn2017synthetic,firpo2018synthetic}, and describe the estimation procedure in Section \ref{eval}. The confidence intervals generally include zero for time periods prior to 1869, when no treatment effect is expected. 

Counterfactual predictions of state government education spending in the absence of the HSA generally tracks the observed treated outcomes until the first treatment time, at which the counterfactual diverges from the increasing observed treated time-series. Taking the mean of the post-treatment impacts, $\hat{\tau}_t^{\text{ATT}}$, the encoder-decoder estimate of the impact of the HSA on the education spending of public land states is 0.681 log points [0.175, 1.186], as reported in the first column of Table \ref{benchmark-compare}. The confidence intervals surrounding this estimate do not contain zero, which indicates that the estimated effect is significantly more extreme than the exact distribution of average placebo effects under the null hypothesis. To put the magnitude of this point estimate in perspective, it represents about 2.5\% of the per-capita total expenditures for public schools in 1929 \citep{snyder2010digest}. Table \ref{benchmark-compare} reports the causal estimates recovered by each of the benchmark estimators used in the placebo tests. The VAR estimated effect is slightly smaller and also statistically significant, whereas the other estimators yield wider confidence intervals that contain zero.
 
\begin{table}[htb]
	\centering
	\begin{adjustbox}{max width=\textwidth}
		\begin{threeparttable}	\footnotesize
			\captionsetup{font=normalsize}
			\caption{ATT estimates by estimator.\label{benchmark-compare}}
			\begin{tabular}{@{}lcc@{}}
	\toprule
	\multicolumn{1}{l}{} &
	\multicolumn{1}{c}{\begin{tabular}[c]{@{}c@{}} $\hat{\tau}^{\text{ATT}}$ \end{tabular}} &
	\multicolumn{1}{c}{\begin{tabular}[c]{@{}c@{}} $\hat{\tau}_{\text{placebo}}^{\text{ATT}}$ \end{tabular}}  \\ \midrule
	\hline \\
	DID 					&  -0.597 [-2.352, 1.351] & 	 -0.030	[-1.212, 1.396]		 \\
	Encoder-decoder (ours) 	&  \bf{0.681 [0.175, 1.186]}   &     0.213 [-1.662, 2.019]		 \\
	LSTM (ours) 			&  0.266 [-0.342, 0.876] &      -0.083 [-2.146, 1.343]		\\
	MC-NNM     				& -0.603 [-2.429, 0.994]  &		0.132 [-0.661, 1.136]		\\
	SCM     				& 0.404 [-0.075, 1.089]   &		0.878 [-0.187, 2.463]		\\
	SCM-L1    				& 0.348 [-0.316, 1.029]   &		0.517 [-0.799, 1.977]\\
	VAR      				& \bf{0.407 [0.231, 1.062]}   &		0.868 [-1.381, 5.110]					\\ \bottomrule
\end{tabular}%
			\begin{tablenotes}[flushleft]
				\small
				\item \emph{Notes:} First column is ATT estimates of the impact of the HSA on log per-capita state government education spending. Second column is placebo ATT estimates, with the placebo initial treatment time set to $T_0/2$. Bracketed values are 95\% randomization confidence intervals. Bold indicates statistical significance at the $\alpha = 0.05$ level. 
			\end{tablenotes}
		\end{threeparttable}
	\end{adjustbox}
\end{table}

\subsection{Sensitivity} \label{imp-sens}

In the main analyses, we impute values in the education spending data that are missing due to lack of coverage by LOCF. Table \ref{benchmark-compare-imp} presents ATT estimates on differently imputed datasets using four alternative imputation methods: $k$-Nearest Neighbor ($k$-NN), linear interpolation, multivariate imputation by chained equations (MICE), and random forests. The encoder-decoder estimates remain significant when missing values are replaced by linear interpolation, and lose their significance when missing values are replaced by the other imputation methods.

We also evaluate the sensitivity of the causal estimates to different configurations of RNNs hyperparameters by varying the hidden activation function (tanh or sigmoid); the number of hidden units per hidden layer (128 or 256 units); early stopping patience (25 or 50 epochs); and the dropout probability ($p=0.2$ or $p=0.5$). We report the ATT estimates by hyperparameter configuration in Table \ref{educ-sens-rnns}. The encoder-decoder causal estimates are positive and significant for half of the 16 different hyperparameter configurations. 

\subsection{Placebo tests} \label{placebo-pretreatment}

We assess the accuracy of the estimators by conducting placebo tests on the pretreatment data, when no treatment effect is expected. Among the actual treated units, we assign treatment times that are equally spaced between the placebo $T_0$, which is half of the actual $T_0$, and $T_0-1$. We then construct randomization confidence intervals for the placebo counterfactual trajectories, which we report in the second column of Table \ref{benchmark-compare}. For each estimator, the confidence intervals contain zero, providing indirect evidence that Assumption \eqref{unconfoundedness} is not violated.

\section{Conclusion} \label{conclusion}

This paper makes a methodological contribution in proposing an RNN-based estimator for estimating the effect of a binary treatment in panel data settings. RNNs are specifically structured to exploit temporal dependencies in the data and can learn nonlinear combinations of control units; the latter is useful when the data-generating process underlying the outcome depends nonlinearly on the history of its inputs. Most real-world time series data have a linear dependency component, for which VARs are suitable, and a nonlinear dependency component that potentially spans across multiple time periods, for which RNNs are suitable. RNNs are unable to handle both linear and nonlinear patterns, which are often both present in real-world time-series, and typically require a large amount of training data to effectively capture nonlinear and potentially long-term dependencies. In placebo tests, we find that RNN-based estimators perform well in terms of minimizing out-of-sample error compared to VAR and other linear estimators on both small- and high dimensional datasets with varying degrees of temporal dependency. An area of further research is extending the method to combining both linear time-series models such as VAR with RNNs in order to more accurately model complex autocorrelation structures in the data \citep[e.g.,][]{DBLP:journals/corr/abs-1709-03159}.

In the empirical application, we estimate the impact of mid-19th century homestead policy on the development of state government public education spending in the U.S. We train RNNs on states unaffected by homestead policy and use the learned weights to predict the counterfactual outcomes of public land states, which were open to homesteading. The encoder-decoder estimate of the impact of homestead policy on the education spending of public land states is 0.681 log points [0.175, 1.186], which represents about 2.5\% of the per-capita total expenditures for public schools in 1929. This estimate is generally robust to the configuration of RNN hyperparameters and the missing data imputation method. The result is consistent with both the historical view that state governments in public land states expanded public education investments to attract eastern migrants, and the political economy view that homesteading deterministically lowered land inequality on the frontier and consequently prevented wealthy landowners from blocking public schooling reforms.

\newpage

\bibliographystyle{rss}
\begin{singlespace}
			\bibliography{references}
\end{singlespace}

\itemize
\end{document}


\maketitle \thispagestyle{empty}
\thispagestyle{empty} 

\begin{abstract}
	\noindent 
The supporting materials include model specifications and implementation details for each of the benchmark estimators used in the placebo test experiments. It also describes the estimation procedure for randomization confidence intervals for average treatment effects used in the application. Lastly, it includes a figure visualizing the extent of missing data in the education spending dataset, and tables reporting the sensitivity of average treatment effects to hyperparameter configuration and missing data imputation method. 
\end{abstract}

\setcounter{figure}{0} \renewcommand{\thefigure}{SM-\arabic{figure}}
\setcounter{table}{0} \renewcommand{\thetable}{SM-\arabic{table}}
\setcounter{section}{0} \renewcommand{\thesection}{SM-\arabic{section}}

\pagenumbering{roman}

\pagebreak
\pagenumbering{arabic}

\section{Benchmark estimators} \label{benchmark-estimators}

The following five estimators are used for comparison with the RNN-based estimators in the placebo tests (Section \ref{placebo}) and main estimates (Section \ref{estimates}).

\subsection{Difference-in-differences (DID)}

The DID model is specified by \citep{athey2018design}:
\begin{equation}
Y_{it}  = \gamma_i + \delta_t + \tau W_{it} + \epsilon_{it},
\end{equation}
where $\gamma_i$ + $\delta_t$ are unit and time fixed effects, respectively. The model is estimated by least squares: 
\begin{equation}
\argmin _{\tau,\gamma_{i},\delta_{t}} \sum_{i=1}^{N} \sum_{t=1}^{T}\left(Y_{i t}-\gamma_{i}-\delta_{t}-\tau W_{i t}\right)^{2}.
\end{equation}
We use the implementation of DID in the \texttt{MCPanel} \textsf{R} package, available at \url{https://github.com/susanathey/MCPanel}.

\subsection{Matrix completion by nuclear norm minimization\\ (MC-NNM)}

The matrix completion model (without covariates) is specified by \citet{athey2017matrix}:
\begin{equation}
	Y_{it} = L_{it} + \gamma_{i} + \delta_{t} + \epsilon_{it}, \label{eq:mc-Y}
\end{equation} 
where $L_{it}$ is a low-rank matrix to be estimated. Estimating $L$ involves minimizing the sum of squared errors via nuclear norm regularized least squares:
\begin{equation}
	\argmin_{L, \boldsymbol{\gamma}, \boldsymbol{\delta}} \Bigg[\frac{1}{|\mathcal{O}|} \sum_{(it) \in \mathcal{O}}  W_{it} \, (1-\mathbf{\hat{\text{e}}}_{it}) + (1-W_{it}) \,\mathbf{\hat{\text{e}}}_{it} \, \bigg(Y_{it} - L_{it} - \gamma_{i} - \delta_{t} \bigg)^2
	+ \lambda_L \norm{L}_\star \Bigg], \label{eq:mc-opt-prop}
\end{equation}
where $\mathcal{O}$ denotes the set of observed $it$ values; i.e., the values for which $W_{it}=0$. The nuclear norm, or sum of singular values, $\norm{\cdot}_\star = \sum_{i} \sigma_i (\cdot)$ is used to yield a low-rank solution for $L$. The value of the penalty hyperparameter $\lambda_L$ is chosen by cross-validation.

In order to place more emphasis on the loss for the values in $\mathcal{O}$ most similar to the missing values in terms of estimated propensity scores, $\hat{e}_{it}$, we use overlap weights so that observed $it$ values under treatment receive a weight of $1-\hat{e}_{it}$ and observed $it$ values under control receive a weight equal to $\hat{e}_{it}$.

We use an extended version of the \texttt{MCPanel} package for implementing MC-NNM with a propensity-weighted loss function, under otherwise default settings. The extended version of the package is available at \url{https://github.com/jvpoulos/MCPanel}.

\subsection{Generalized synthetic control method (SCM)}

The SCM \citep{abadie2010synthetic} compares a single treated unit with a synthetic control that combines the outcomes of multiple control units on the basis of their pre-intervention similarity with the treated unit. 

\citet{doudchenko2016balancing} and \cite{athey2017matrix} show that the SCM can be interpreted as regressing the pre-treatment outcomes of a single treated unit on the control unit outcomes during the same periods. The parameters estimated on the controls are then used to predict the counterfactual outcomes for a single treated unit, $i=0$:
\begin{align}\label{adh} \nonumber
\hat{Y}_{0t} &= \sum_{i=1}^{\textrm{N}} \hat{\omega}_i Y_{it}, \qquad \forall \, t= \textrm{T}_0, \ldots, \mathrm{T},\\
\textrm{where } \hat{\omega} &=\argmin_{\omega}\sum_{s=1}^{\textrm{T}_0} \left(Y_{0s}-\sum_{i=1}^{\textrm{N}} \omega_i Y_{0s}\right)^2, \qquad \textrm{s.t. } w_i \geq 0, \sum w_i =1.
\end{align}
A separate model is subsequently fit for each $i, \ldots, N_t$ treated units. Note that Eq.~\eqref{adh} imposes the restrictions of the original SCM, namely zero intercept and non-negative regression weights that sum to one. 

We use the \texttt{MCPanel} implementation with default settings, except we limit gradient values within [-5,5] in order to facilitate convergence. The underlying algorithm for estimating the parameters is exponentiated gradient descent \citep{kivinen1997exponentiated}. 

\subsection{Generalized synthetic control method with lasso\\ (SCM-L1)}

The SCM-L1 \citep{doudchenko2016balancing,athey2017matrix} relaxes the zero-intercept and weight restrictions of Eq.~\eqref{adh}. The counterfactual outcomes for treated unit $i=0$ is:
\begin{align}\label{vt} \nonumber
\hat{Y}_{0t} &= \hat{\mu} + \sum_{i=1}^{\textrm{N}} \hat{\omega}_i Y_{it}, \qquad \forall \, t= \textrm{T}_0, \ldots, \mathrm{T},\\
\textrm{where } \left(\hat{\mu}, \hat{\omega}\right) &=\argmin_{\mu, \omega}\sum_{s=1}^{\textrm{T}_0} \left(Y_{0s}-
\mu-\sum_{i=1}^{\textrm{N}} \omega_i Y_{0s}\right)^2, 
\end{align}
where a separate model is fit for each $i, \ldots, N_t$ treated units. Intuitively, the generalized SCM is a convex combination of control units with intercept $\mu$ and weight $\omega_i$ for control units $i, \ldots, \textrm{N}$. The model is fit with $N+1$ predictors, including the number of control units and the intercept, and $\textrm{T}_0$ observations. Eq.~\eqref{vt} is estimated by lasso regression \citep{tibshirani1996regression,tibshirani2012strong} in order to reduce the relative size of the predictor set. We use the \texttt{MCPanel} implementation with default settings.

\subsection{Vector autoregression (VAR)}

For a vector of outputs $\textbf{y}_{t} = \left(y_t[1], y_t[2], \ldots y_t[N]\right)^\top$, we fit a stationary VAR with a single lag on the control outcomes:
\begin{equation}\label{var}
\textbf{y}_{t+1} = \beta \textbf{y}_{t} + \epsilon_{t+1}, \, \qquad \forall y_t[i] \text{ with } W_i = 0.
\end{equation}
The coefficient matrix $\beta$ is estimated by lasso regression in order to reduce the size of the predictor set \citep{callot2014oracle,kock2015oracle}. VAR forecasts on the treated outcomes are made by
\begin{equation}\label{var-pred}
\hat{\textbf{y}}_{t} = \hat{\beta} \textbf{y}_{t}, \, \qquad \forall y_t[i] \text{ with } W_i = 1.
\end{equation}
Similar to the RNNs training procedure, we fit the VAR on control outcomes using a sliding window of size $T_0$ with step size of one. We use the authors' \textsf{R} package \texttt{lassovar} with default settings for implementing VAR.

\section{Randomization confidence intervals} \label{eval}

Under the assumption that treatment has a constant additive effect, $\Delta$, we construct a two-sided 95\% randomization confidence interval for the causal estimand $\tau^{\textup{ATT}}$ defined by Eq.~\eqref{eq:att-avg} by inverting the randomization test (Algorithm \ref{alg:randomization-ci}). 
\begin{algorithm}[H]
	\begin{algorithmic}[1]
		\item[ ] {\textbf{inputs:}} (a) post-treatment predictions and observed values for control units, or $\widehat{{Y}_{it}}(1) \text { and } {Y}_{it}(0) \text { for } W_{i}=0 \text { and } t \geq T_{0}$, respectively; (b) estimated ATT averaged over the post-treatment period, $\hat{\tau}^{\textup{ATT}}$; (c) number of control units, $\text{J}$
		\item[ ] {\textbf{output:}} 95\% randomization confidence interval for $\hat{\tau}_t^{\textrm{ATT}}$
		\FOR{$l=1$ to $\mathcal{L}$} 
		\STATE Randomly sample without replacement constant treatment effect $\Delta_i$ \label{endpoints}
		\STATE Subtract $\Delta_i$ from $\hat{\tau}^{\textup{ATT}}$, resulting in test statistic $\delta_{\Delta_i}$
		\FOR{$q=1$ to $\mathcal{Q}$}
		\STATE Randomly sample without replacement which $\text{J}-1$ control units are assumed to be treated \label{placebo-treated}
		\STATE Calculate average placebo treated effect $\hat{\tau}_{iq}^{\text{ATT}}$
		\STATE Count the number of $\hat{\tau}_{iq}^{\textrm{ATT}}$  that are greater than or equal to $|\delta_{\Delta_i}|$  \label{counts}
		\ENDFOR
		\STATE Divide count obtained from Step \ref{counts} by $\mathcal{Q}$ to estimate two-sided $p$ values, $\hat{p}$
		\STATE Collect all values of $\delta_{\Delta_i}$ that yield $\hat{p}$ greater than $\alpha=0.05$
		\ENDFOR
	\end{algorithmic}
		\caption{Randomization confidence intervals}
\label{alg:randomization-ci}
\end{algorithm}
In Step \ref{endpoints}, we find the endpoints of the confidence interval by randomly sampling without replacement $\mathcal{L} =5,000$ values of $\Delta$. In Step \ref{placebo-treated}, there are $\mathcal{Q} = \sum\limits_{\text{q}=1}^{\text{J}-1} {\text{J} \choose \text{q}}$ possible average placebo effects, where $\text{J}$ is the number of control units. Since calculating $\mathcal{Q}$ can be computationally burdensome for relatively high values of $J$, we set $\mathcal{Q} = 10,000$.

\section{Tables \& Figures}

\begin{figure}[htbp]
	\centering
		\includegraphics[width=\textwidth]{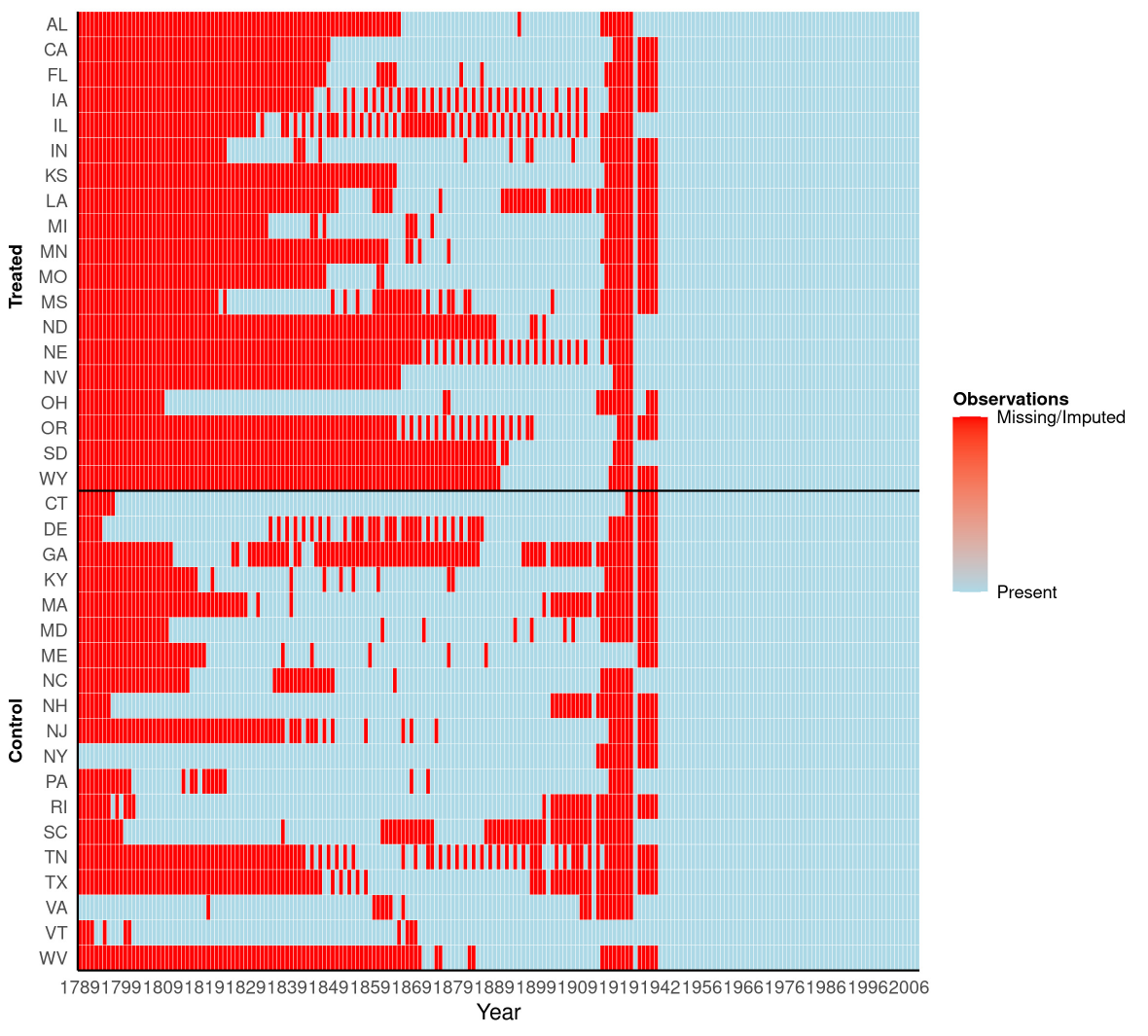}
	\caption{State-year observations in the education spending dataset that are missing and to be imputed. `Control' are state land states that comprise the training set and `Treated' are public land states that comprise the test set. \label{fig:missing-heatmap}}
\end{figure}

Figure \ref{fig:missing-heatmap} visualizes the extent of the missing values in the education spending data, where 34.1\% of values in the dataset are missing (25.5\% and 42.7\% missing in the training and test sets, respectively) and need to be imputed. In the placebo test experiments (Section \ref{placebo}) and in the main analysis (Section \ref{estimates}), the missing values are imputed by LOCF, with remaining missing values carried backward. We implement LOCF with the \texttt{imputeTS} \textsf{R} package \citep{moritz2017imputets} with default settings. 

Table \ref{benchmark-compare-imp} presents ATT estimates on differently imputed datasets using four alternative imputation methods (implementation details provided below): $k$-Nearest Neighbor ($k$-NN), linear interpolation, multivariate imputation by chained equations (MICE), and random forests. Each imputation method is performed separately on the training and test sets to avoid data contamination. Lastly, Table \ref{educ-sens-rnns} assesses the sensitivity of the causal estimates in the main analysis (on LOCF imputed data) to different configurations of RNNs hyperparameters. 

\begin{description}
	\item[$k$-NN] imputation using the median value of $k=5$ nearest neighbors based on a variation of Gower distance \citep{gower1971general}; implemented with the \texttt{VIM} \textsf{R} package \citep{kowarik2016imputation} with default settings.
	\item[Linear interpolation] uses linear interpolation to replace missing values; implemented with the \texttt{imputeTS} package with default settings.
	\item[MICE] multiple imputation approached via Fully Conditional Specification (FCS); implemented with the \texttt{mice} \textsf{R} package \citep{buuren2010mice} with default settings.
	\item[Random forests] iterative imputation method based on random forests; implemented with the \texttt{missForest} \textsf{R} package \citep{stekhoven2012missforest} with default settings.
\end{description}

\begin{table}[p]
	\centering
	\begin{adjustbox}{max width=\textwidth}
		\begin{threeparttable}
			\captionsetup{font=normalsize}
			\caption{ATT estimates by estimator and imputation method.\label{benchmark-compare-imp}}
			\begin{tabular}{@{}lccccc@{}}
	\toprule
	\multicolumn{1}{l}{\emph{Imputation method:}} &
		\multicolumn{1}{c}{\begin{tabular}[c]{@{}c@{}} $k$-NN \end{tabular}} &
	\multicolumn{1}{c}{\begin{tabular}[c]{@{}c@{}} Linear\\interpolation \end{tabular}} &
	 \multicolumn{1}{c}{\begin{tabular}[c]{@{}c@{}} MICE \end{tabular}} &
	 \multicolumn{1}{c}{\begin{tabular}[c]{@{}c@{}} Random\\ forests \end{tabular}} \\ \midrule
	\hline \\
	DID 					& -1.275 [-4.100, 0.281]	& -0.612 [-2.388, 1.399]			& -2.088 [-5.812, 1.601]		& -1.230 [-3.700, 0.735]	\\
	Encoder-decoder (ours)  &	-0.302 [-1.161, 0.555]	& \bf{0.809 [0.110, 1.506]}				&	-0.025 [-1.370, 1.318]	& 0.226 [-0.628, 1.082] ]	 						 \\
	LSTM (ours)				&	-0.398 [-1.361, 0.542]	& 0.380 [-0.362, 1.125]				&	-0.224 [-1.615, 1.171]		& 0.145	[-0.762, 1.051]  	 						\\
	MC-NNM     				& -1.104 [-3.629, 1.178]	& -0.608 [-2.433, 1.127]			& -1.800 [-5.591, 1.985]	& -1.062 [-3.449, 1.002] \\
	SCM     				& 0.577	[-0.207, 1.511]		& 0.388	[-0.125, 1.050]				& 0.407 [-0.671, 1.500]			& 0.236 [-0.481, 0.997]	\\
	SCM-L1    				& 0.394 [-0.510, 1.345]		& 0.330	[-0.266, 0.999]				& 0.456 [-2.262, 3.396]			& 0.293 [-0.427, 1.049]	\\
	VAR      				& \bf{0.598 [0.043, 1.576]}		& 0.319	[-0.034, 0.874]		 	& \bf{1.091 [0.454, 2.525]}	    & 0.549 [-0.010, 1.178]	\\ \bottomrule
\end{tabular}%
			\begin{tablenotes}[flushleft]
				\small
				\item \emph{Notes:} ATT estimates on datasets with differently imputed missing values. Bracketed values are 95\% randomization confidence intervals for $\hat{\tau}^{\text{ATT}}$. Bold indicates statistical significance at the $\alpha = 0.05$ level. 
			\end{tablenotes}
		\end{threeparttable}
	\end{adjustbox}
\end{table}

\begin{table}[htbp]
	\centering
	\begin{adjustbox}{max width=\textwidth}
		\begin{threeparttable}	\footnotesize
			\captionsetup{font=normalsize}
			\caption{ATT estimates by RNNs hyperparameter configuration.\label{educ-sens-rnns}}
			\begin{tabular}{@{}lccccc@{}}
	\toprule
	\multicolumn{4}{c}{} &
	\multicolumn{1}{c}{\begin{tabular}[c]{@{}c@{}}Encoder-decoder \end{tabular}} &
	\multicolumn{1}{c}{\begin{tabular}[c]{@{}c@{}}LSTM \end{tabular}} \\ \midrule
	\hline \\
	\emph{Hidden} & \emph{Hidden} & \emph{Patience} & \emph{Dropout} 	&			&	\\
	\emph{activation}	 & \emph{units} & \emph{(epochs)} & \emph{($p$)} 	&			&	\\
			 &  & & 	&			&	\\
	sigmoid & 256 &  50 & 0.5 		& 0.361 [-0.337, 1.059]		&  	0.292 [-0.426, 1.012]	 \\
	tanh & 256 & 50 & 0.5			& 0.524 [-0.017, 1.069]		& 	0.388 [-0.186, 0.961]		\\
	sigmoid & 128 &  50 & 0.5 		& 0.458 [-0.235, 1.149]		&  0.295 [-0.439, 1.043]	  \\
	tanh & 128 & 50 & 0.5			& \bf{0.661 [0.171, 1.153]} & 0.277 [-0.324, 0.880] 	  \\
	sigmoid & 256 &  25 & 0.5 		& 0.453 [-0.238, 1.145]		&  0.246 [-0.502, 1.005]	 \\
	tanh & 256 & 25 & 0.5			& \bf{0.553 [0.017, 1.087]}	& 0.329 [-0.250, 0.910] 	\\
	sigmoid & 128 &  25 & 0.5 		& 0.427 [-0.291, 1.145]		&  0.307 [-0.442, 1.065]	\\
	tanh & 128 & 25 & 0.5& \bf{0.960 [0.403, 1.514]} & 0.400 [-0.187, 0.987]		\\
	sigmoid & 256 &  50 & 0.2 		& \bf{0.643 [0.034, 1.257]}	&  0.177 [-0.459, 0.813]	\\
	tanh & 256 & 50 & 0.2			& \bf{0.828 [0.442, 1.214]}	& \bf{0.606 [0.174, 1.037]}		\\
	sigmoid & 128 &  50 & 0.2 		& 0.325 [-0.358, 1.015]		& 0.193 [-0.489, 0.887] 	\\
	tanh & 128 & 50 & 0.2			& \bf{0.858 [0.404, 1.311]}	& 0.279 [-0.195, 0.752]		\\
	sigmoid & 256 &  25 & 0.2 		& 0.340 [-0.311, 1.001]		& 0.150 [-0.559, 0.871]	 \\
	tanh & 256 & 25 & 0.2			& \bf{0.745 [0.286, 1.206]}	& 0.329 [-0.114, 0.771]		\\
	sigmoid & 128 &  25 & 0.2 		& 0.359 [-0.321, 1.060]		& 0.197 [-0.483, 0.888] 	\\
	tanh & 128 & 25 & 0.2			& \bf{0.856 [0.404, 1.309]} 	& 0.466 [-0.049, 0.983]		\\	
\bottomrule
\end{tabular}%
			\begin{tablenotes}[flushleft]
				\small
				\item \emph{Notes:} estimates of the impact of the HSA on log per-capita state government education spending (1942\$). Bracketed values are 95\% randomization confidence intervals for $\hat{\tau}^{\text{ATT}}$. Bold indicates statistical significance at the $\alpha = 0.05$ level. 
			\end{tablenotes}
		\end{threeparttable}
	\end{adjustbox}
\end{table}

\bibliographystyle{rss}
\begin{singlespace}
\bibliography{references}
\end{singlespace}

\itemize